\documentclass[10pt,twocolumn,letterpaper]{article}

\usepackage{cvpr}
\usepackage{times}
\usepackage{epsfig}
\usepackage{graphicx}
\usepackage{amsmath}
\usepackage{amssymb}
\usepackage{booktabs}
\usepackage{caption}
\usepackage{cuted}
\usepackage{capt-of}
\usepackage{enumitem}

\usepackage[pagebackref=true,breaklinks=true,letterpaper=true,colorlinks,bookmarks=false]{hyperref}

\usepackage[usenames, dvipsnames]{color}

\cvprfinalcopy 

\def\cvprPaperID{2471} 

\ifcvprfinal\fi
\begin{document}

\title{Hierarchical Deep Stereo Matching on High-resolution Images}

\author{Gengshan Yang$^{1\thanks{Work done while interning at Argo AI. Code available \href{www.contrib.andrew.cmu.edu\/\~gengshay\/cvpr19stereo}{here}.}}$,\quad Joshua Manela$^2$,\quad Michael Happold$^2$,\quad Deva Ramanan$^{1,2}$\\
$^1$Carnegie Mellon University,\quad $^2$Argo AI\\
{\tt\small gengshany@cmu.edu}\quad {\tt\small \{jmanela,\,mhappold,\,dramanan\}@argo.ai}\\
\vspace*{-7in}
}

\maketitle 
\begin{strip}\centering
\includegraphics[trim={0cm 0.5cm 0cm 0cm},clip,width=\textwidth]{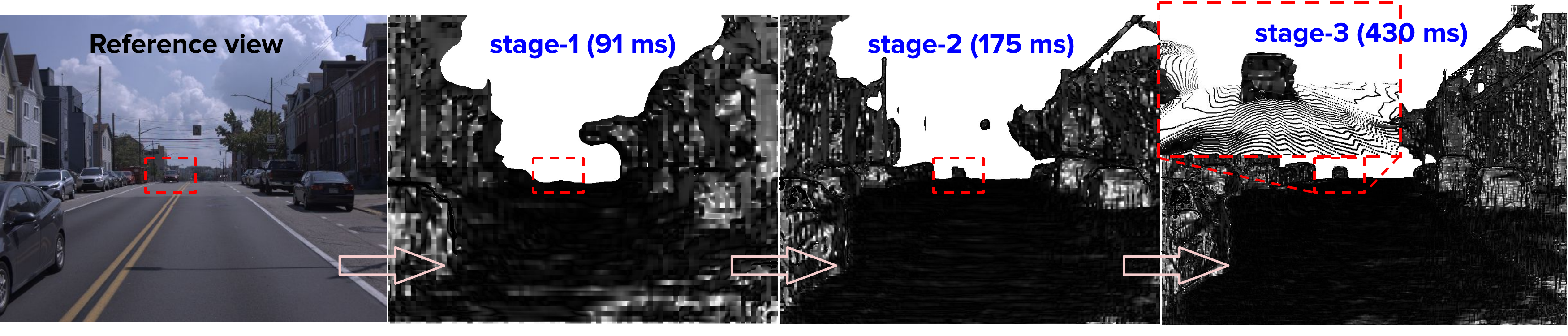}
\captionof{figure}{Illustration of on-demand depth sensing with a coarse-to-fine hierarchy on the proposed dataset. Our method (HSM) captures the coarse layout of the scene in 91 milliseconds, finds the far-away car (shown in the red box) in 175 ms, and recovers the details of the car given extra 255 ms.
\label{fig:feature-graphic}}
\end{strip}
    

\begin{abstract}
We explore the problem of real-time stereo matching on high-res imagery. 
Many state-of-the-art (SOTA) methods struggle to process high-res imagery because of memory constraints or speed limitations. To address this issue, we propose an end-to-end framework that searches for correspondences incrementally over a coarse-to-fine hierarchy.  Because high-res stereo datasets are relatively rare, we introduce a dataset with high-res stereo pairs for both training and evaluation. Our approach achieved SOTA performance on Middlebury-v3 and KITTI-15 while running significantly faster than its competitors. The hierarchical design also naturally allows for anytime on-demand reports of disparity by capping intermediate coarse results, allowing us to accurately predict disparity for near-range structures with low latency (30ms). 
We demonstrate that the performance-vs-speed tradeoff afforded by on-demand hierarchies may address sensing needs for time-critical applications such as autonomous driving.
\end{abstract}


\section{Introduction}

In safety-critical applications such as autonomous driving, it is important to accurately perceive the depth of obstacles with low latency. Towards that end, we explore the problem of real-time stereo matching on high-res imagery.

{\bf LiDAR vs Stereo:} LiDAR is the common choice for outdoor depth sensing~\cite{strecha2008benchmarking}. However, they are fundamentally limited in spatial density, particularly for long-range sensing.  Only so many beams and detectors can be packed together before cross-talk occurs. In principle, one can increase density by scanning slowly, but this introduces latency and rolling shutter effects that can be devastating for dynamic scenes. Density of sensor measurements is crucial for recognizing and segmenting objects at range. High-res, global shutter stereo has the potential to address these limitations.

{\bf Why high-res?} It is widely recognized that stereo is unreliable for long-range depth sensing \cite{pinggera2014know}: 
the metric error in estimated depth $\Delta Z$ of a triangulation-based stereo system with baseline $b$, focal length $f$, and pixel matching error $\Delta d$ can be written as $\Delta Z = Z^2 \frac{\Delta d}{b\cdot f}$. Hence depth error increases quadratically with depth, implying that stereo will provide unstable far-field depth estimates. But this is important for navigation at even moderate speeds - (see the supplement for stopping distance profiles). 
Though one can attempt to tweak the other factors to reduce the error, 
higher-resolution (larger $f$) appear to be the most promising avenue: innovations in CMOS and CCD sensor technology have allowed for mass-market, low-cost solutions for high-resolution cameras.

\begin{figure*}[!ht]
\centering
\includegraphics[width=\linewidth]{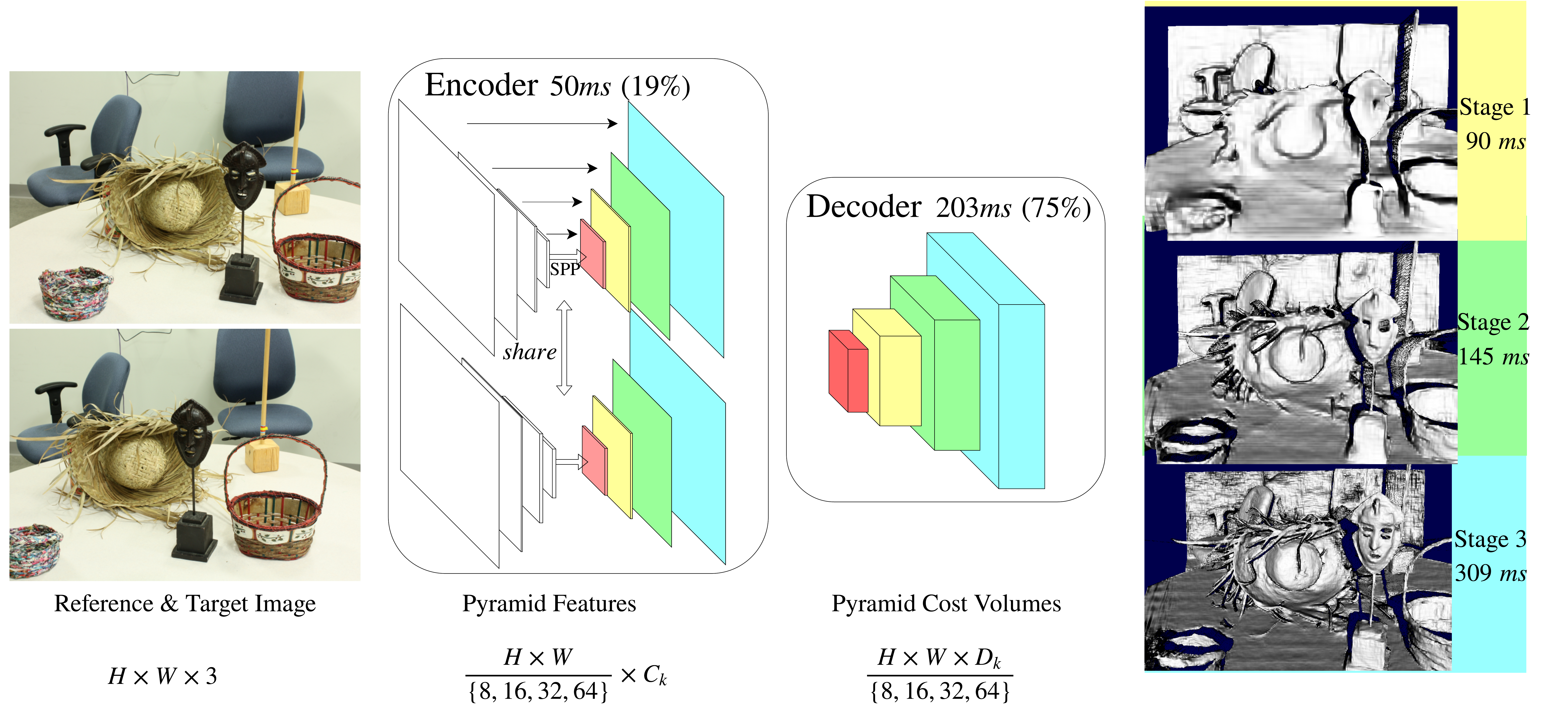}
\caption{Our on-demand, low-memory architecture for high-res stereo. Given a rectified pair of high-res images, we compute multiscale descriptors for each with a custom resnet ``butterfly" encoder-decoder network (which we refer to as our pyramid {\em encoder}). These descriptors are used to construct 4D feature volumes at each scale ($\frac{C_k\times H\times W \times D}{\{8,16,32,64\}}$, where scale $k\in\{1,2,3,4\}$ and $C_k \in \{16,16,16,32\}$), by taking the difference of potentially matching features extracted from epipolar scanlines. 
Each feature volume is {\em decoded} or filtered with 3D convolutions, making use of striding along the disparity dimensions to minimize memory. The decoded output is (a) used to predict 3D cost volumes that generates on-demand disparity estimates for the given scale and (b) upsampled so that it can be combined with the next feature volume in the pyramid.
 and $D_k \in \{\frac{D_{max}}{4},\frac{D_{max}}{2},D_{max},D_{max}\}$ represent the number of feature channel and the number of disparity bins in the $k$th scale, and the time is measured for $6$-megapixel input with a disparity search range of $256$px on Titan X Pascal. 
}
\label{fig:archnew}
\end{figure*}

{\bf Challenges:}
Although high-res stereo matching is desirable, there are several practical challenges: optimization-based stereo methods are accurate but cannot scale to high-resolution with regard to both running time and memory overhead. When applied to down-scaled images, these methods run faster, but gives blurry results and inaccurate disparity estimates for the far-field. Recent ``deep" stereo methods perform well on low-resolution benchmarks~\cite{chang2018pyramid,ilg2018occlusions,Liang2018Learning,Menze2015CVPR,yang2018segstereo}, while failing to produce SOTA results on high-res benchmarks~\cite{scharstein2014high}. This is likely due to: 1) Their architectures are not efficiently designed to operate on high-resolution images. 2) They do not have enough high-resolution training data.

{\bf Approach:} We propose an end-to-end framework that efficiently searches for correspondences over hierarchies. Our model reasons in a coarse-to-fine-manner, inspired by classic work on correspondence estimation in stereo and optical flow~\cite{anandan1989computational,lucas1981iterative,yang2006real}. Coarse resolution images are used to estimate large disparities, which are then used to bias/pre-warp fine-scale disparity estimates. Though quite efficient, coarse-to-fine methods struggle to match thin structures that ``disappear" at coarse resolutions~\cite{brox2011large}. Instead, our model computes a {\em high-res} encoder feature that is processed with coarse-to-fine (decoder) feature volumes that gradually increase in resolution. 
Crucially, the initial coarse volume can generate rough estimates of large-disparity objects {\em before} the full pipeline is finished. This allows our network to generate reports of closeby objects on-demand, which can be crucial for real-time navigation at speed. 

{\bf Data:} High-res stereo efforts suffer from the lack of benchmark data, for both training and evaluation. We have collected two datasets of high-res rectified stereo pairs, consisting of real-data from an autonomous vehicle and synthetic data from an urban simulator. Interestingly, we show that synthetic data is a valuable tool for training deep stereo networks, particularly for high-res disparity estimation. Real-world calibration and rectification become challenging at high-res, and introducing realistic calibration errors through data-augmentation 
is helpful for training.

Our main contributions are as follows:
\begin{enumerate}[topsep=1ex,itemsep=1ex,partopsep=1ex,parsep=1ex]
    \item We propose a hierarchical stereo matching architecture that processes high-resolution images while being able to perform on-demand computation in real-time.
    \item We synthesize a high-res stereo dataset for training and collect a real high-res dataset for testing.
    \item We introduce a set of stereo augmentation techniques to improve the model's robustness to calibration errors, exposure changes, and camera occlusions.
    \item We achieve SOTA accuracy on Middlebury and KITTI while running significantly faster than the prior art.
\end{enumerate}

\section{Related Work}


Stereo matching is a classic task in computer vision~\cite{scharstein2002taxonomy}. Traditional methods take rectified image pairs as input (though extensions to multiview stereo exist~\cite{strecha2008benchmarking}), extract local descriptors at candidate patches~\cite{tola2008fast,tola2010daisy} and then build up $3$-dimensional cost volumes across corresponding epipolar scanlines~\cite{ohta1985stereo}. To ensure global ordering and consistency constraints, global optimization techniques are applied with a considerable cost of time and memory, which also poses limitations on scale~\cite{hosni2013fast,kolmogorov2001computing}.

{\bf Efficient high-resolution stereo: } To mitigate this problem, SGM~\cite{hirschmuller2008stereo} and ELAS~\cite{geiger2010efficient} describe efficient matching algorithms that allow for ~$3$FPS on $1.5$-megapixel images with a strong performance. However, both methods struggle to scale to $6$-megapixel images: for example, SGM requires $16$ GB to process a disparity search range of 700px, and surprisingly, performance {\em drops} when compared to a variant that processes lower-res inputs ~\cite{legendre2017high}. While other work has also explored efficient high-res processing \cite{legendre2017high,sinha2014efficient}, they do not appear to meet the accuracy and speed requirements for real-time sensing in autonomous driving.


{\bf Deep stereo matching:} Deep networks tuned for stereo estimation can take advantage of large-scale annotated data. They now produce SOTA performance on several stereo benchmarks, though with considerable use of memory and time.  Zbontar et al. and Luo et al.~\cite{luo2016efficient,zbontar2016stereo} use siamese networks to extract patch-wise features, which are then processed in a traditional cost volume by classic post-processing. Some recent methods \cite{chang2018pyramid,kendall2017end,khamis2018stereonet,Liang2018Learning, MIFDB16, pang2017cascade}
 replace post-processing with 2D/3D convolutions applied to the cost volume, producing SOTA performance on the KITTI benchmark. However surprisingly, none of them outperform traditional methods on Middlebury, possibly due to 1) memory constraints and 2) lack of training data. 
 Although one may run low-res models on crops of high-res images and stitch together the predictions, however, challenges are that: 1) crop borders can be challenging to match; 2) contextual information may not be well-utilized; and 3) most importantly, this dramatically increases run-time latency. 
 
 To our knowledge, we are the first to successfully address these issues and apply deep networks on high-res stereo matching: We propose an efficient hierarchical stereo matching architecture to address the efficiency issue, and leverage high-res synthetic data as well as novel augmentation techniques to overcome data scarcity.

\begin{figure}
    \centering
    \includegraphics[width=\linewidth]{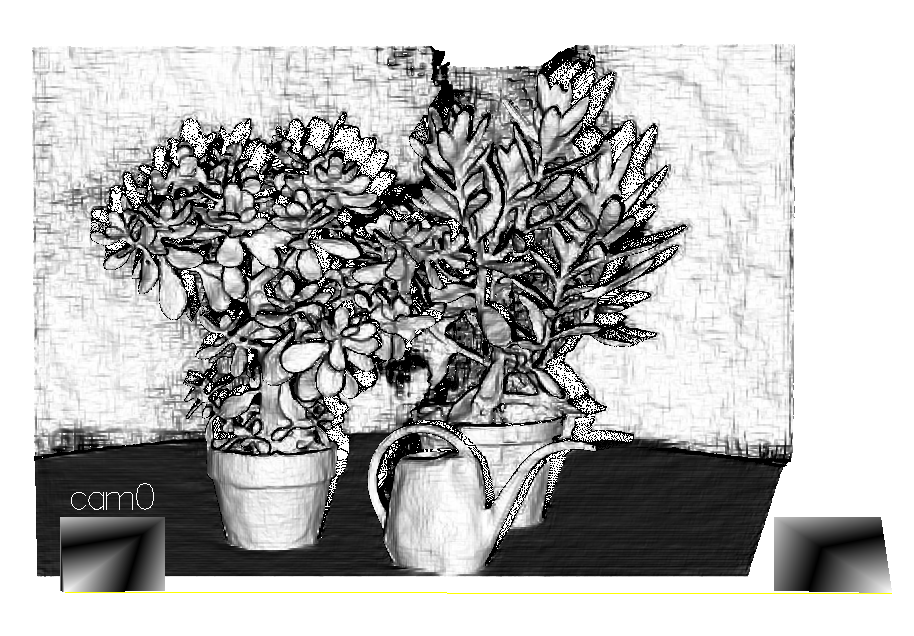}
    \caption{Reconstruction of Middlebury test scene ``Plants''. More results can be found at the project website.}
    \label{fig:middlebury-vis-plants}
\end{figure}

{\bf Coarse-to-fine CNNs:}
Coarse-to-fine design in CNNs dates back to FCN and U-Net \cite{long2015fully,ronneberger2015u}, which leverages multi-scale features and aggregate coarse-to-fine predictions to improve semantic segmentation. DispNet-based architectures \cite{MIFDB16,pang2017cascade} adopts an encoder-decoder scheme with skip-connections, which compute multi-scale cost volumes by correlation and apply 2D convolution/up-convolutions to refine predictions from coarse-to-fine. Recently, PWCNet \cite{sun2018pwc} uses a coarse-to-fine architecture to warp features and achieves SOTA results in optical flow estimation. Closest to our approach, GCNet~\cite{kendall2017end} constructs hierarchical 4D feature volumes and processes them from coarse to fine using 3D convolutions, but we differ in our successful application of coarse-to-fine principles to high-resolution inputs and anytime, on-demand processing. 
\section{Method}
In this section, we describe the key ingredients of our approach: 1) an efficient hierarchical stereo matching architecture, 2) a set of novel asymmetric augmentation techniques, and 3) a high-resolution synthetic dataset for training. We also introduce a high-resolution stereo benchmark for real-world autonomous driving.
\subsection{Hierarchical Stereo Matching (HSM) network}
Our core idea of designing the hierarchical coarse-to-fine network is to first aggressively downsample high res images through the network while extracting multi-scale features, and then use potential correspondences to gradually build up a pyramid of cost volumes that increase in resolution. We provide precise layer and filter dimensions in the supplement.

{\bf Design principles:} We find that coarse-to-fine design principles are crucial, specifically making use of 1) spatial pyramid pooling (SPP)~\cite{zhao2017pyramid}, which allows features to dramatically increase in receptive field. Without this, features tend to have too small a receptive field compared to the rest of the high-res image. The original implementation in SPP~\cite{zhao2017pyramid} upsampled pyramid features back to the original resolution. To reduce memory, we keep pooled features in their native coarse resolution;  2) 3D convolutions that are strided in the disparity dimension, allowing us to process high-res cost volumes efficiently; 3) multi-scale loss functions. The network architecture is shown in Figure \ref{fig:archnew}.


\begin{figure}
\includegraphics[width=\linewidth]{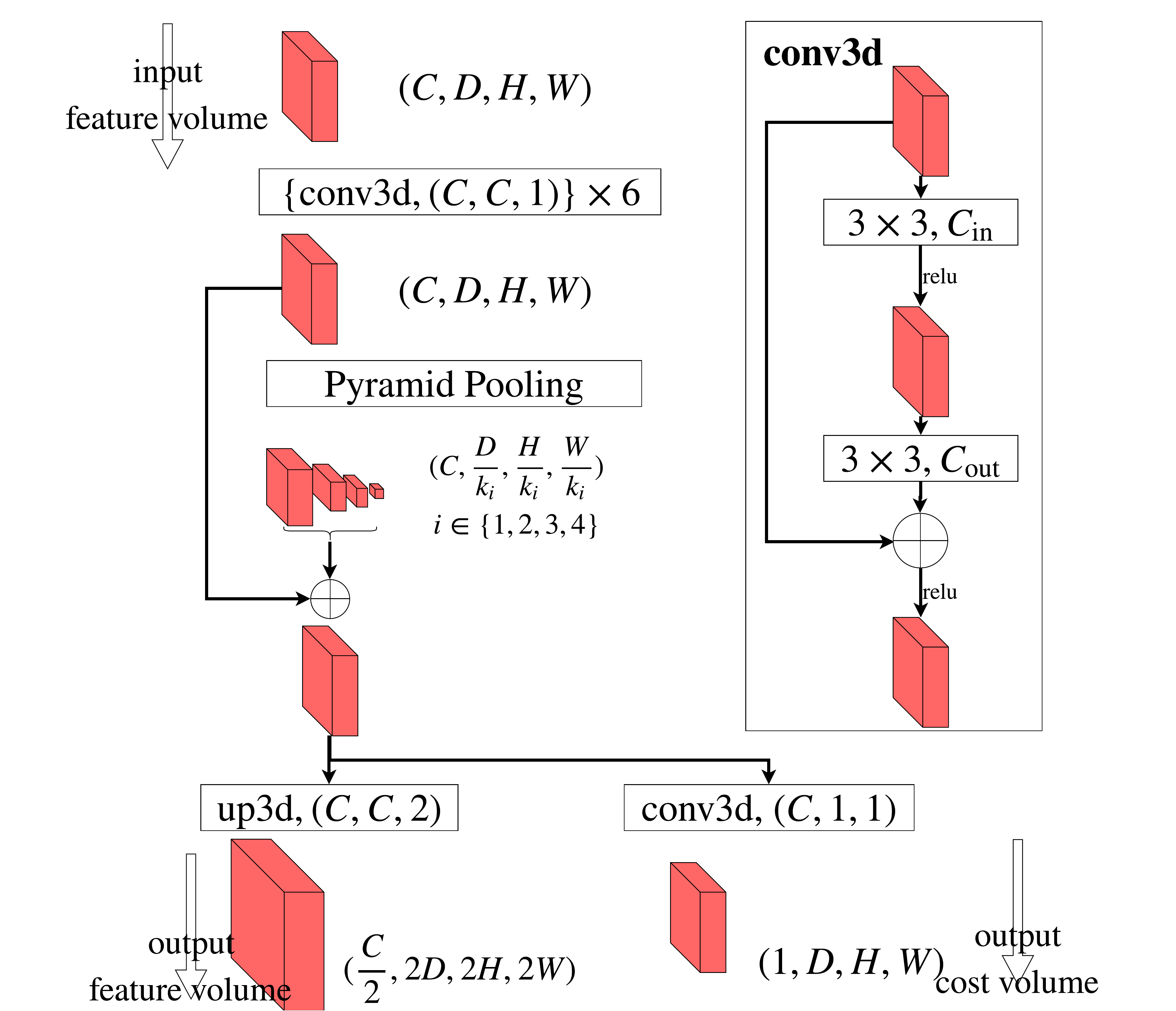}
\caption{Hierarchical feature volume decoder. 3D Convolutions are defined by (in\_channels,out\_channels,stride) and feature volumes are defined by (channels, disparity\_channels, height, width). To reduce memory constraints, we make use of strided disparity channels for the last and second-to-last volume in the pyramid. }
\label{fig:decoder}
\end{figure}

{\bf Feature pyramid encoder:}
We use a feature pyramid encoder to extract descriptors for coarse-to-fine matching. To efficiently extract features with different levels of details while maintaining the coarse-scale information, we adopt an encoder-decoder architecture with skip connections. Our feature encoder consists of custom resnet backbone with 4 residual blocks, followed by 4 SPP layers (again, to increase the receptive fields with limited computation and memory). 

{\bf Feature volumes:}
We obtain such features for both left and right image, and then construct a $4$D {\bf feature volume}~\cite{chang2018pyramid,kendall2017end,khamis2018stereonet} by considering differences between pairs of potentially-matching descriptors along horizontal scanlines. We construct a pyramid of $4$ volumes, each with increasing spatial resolution and increasing disparity resolution. While cost volumes are traditionally $3$D (height H by width W by disparity D), our feature volume includes a $4$th dimension representing the number of feature channels C, which increases for later layers in the encoder. 

{\bf Feature volume decoder:}
Figure \ref{fig:decoder} visualizes the {\em decoding}, or filtering, of each feature volume. Let us first define a conv3D ``block" as two 3D convolutions with a residual connection. 1) The feature volume is filtered by $6$ conv3D blocks. 
2) As was the case for feature extraction, we then apply Volumetric Pyramid Pooling (our extension of SPP to feature volumes) to generate features that capture sufficient global context for high-res inputs. 3a) The output is trilinearly-upsampled to a higher spatial (and disparity) resolution so that it can be fused with the next 4D feature volume in the pyramid. 3b) To report on-demand disparities computed from the current scale, the output is processed with another conv3D block to generate a 3D output cost volume. This cost volume can directly report disparities {\bf before} subsequent feature volumes downstream in the pyramid are ever computed. 

{\bf Multi-scale loss:}
We train the network to make predictions at different scales in the training phase, which allows for {\em on-demand} disparity output at any pyramid level, and also serves to regularize the overall network: 
\begin{align*}
L=L_1+\frac{1}{2^2}L_2+\frac{1}{2^4}L_3+\frac{1}{2^6}L_4
\end{align*}
where losses are scaled to account for the increased disparity resolution at each pyramid level. $L_1$ represents the loss on the finest level, and $L_4$ is the loss on the most coarse level. A natural loss is a softmax distribution over candidate disparities at the current pyramid level. We found that expected disparity, as in GCNet~\cite{kendall2017end}, worked better. 



\subsection{Stereo data augmentation}
In order to train our network, we find it crucial to make use of high-res training data and specific strategies for data augmentation. We discuss both below. Most conventional stereo systems make several assumptions on the target and reference view image pairs, including 1) both images are under the same imaging condition, 2) cameras are perfectly calibrated and 3) there is no occlusion and each pixel can find a match. However, these assumptions do not always hold in real-world scenarios. We propose $3$ asymmetric augmentation techniques to tackle these issues accordingly for our learning-based approach.

{\bf y-disparity augmentation:} Most stereo systems assume that cameras are perfectly calibrated and correspondences lie on the same horizontal scan-line. However, it is difficult to perfectly calibrate a high-res image pair, particularly during large temperature changes and vibrations~\cite{hirschmuller2009stereo}. Such errors result in ground-truth disparity matches that have a $y$ component (e.g., match to a different horizontal scanline). 
Hierarchical matching partially mitigates this issue by biasing matching at the coarse scale, where calibration error is not as severe. 
Another approach is to force the network to {\em learn} robustness to such errors in training time. 
Notice that errors in camera calibration can be represented by a homography $H\in \mathbb{R}^{3\times3}$, resulting in $I_{imperfect}({\bf x})=I_{perfect}(\omega({\bf x},H{\bf x}))$ where ${\bf x}$ is the image coordinate. To mimic the real-world calibration error, we warp the target view image according to the calibration error matrix. To further constrain the space of the warped images, we limit $H$ to be a rigid 2D transformation. 

{\bf Asymmetric chromatic augmentation:} It is inevitable that stereo cameras may under different lighting and exposure conditions, for example, one camera is under the shadow. Therefore making algorithms robust to such imaging asymmetry is critical out of safety concern. We achieve this goal by applying different chromatic augmentation to both reference and target images, in the hope that our feature encoding network may learn a representation robust to such imaging variants.

{\bf Asymmetric masking:}
Most stereo matching algorithms assume that correspondences always exist in the target view. However, this assumption does not hold when occlusion occurs or correspondences are difficult to find. On the other hand, monocular cues, such as shape and continuity, as well as contextual information are found helpful in estimating the disparity. To force the model to rely more on contextual cues, we apply asymmetric masking, which randomly replaces a rectangular region in the target view with mean RGB values of the whole image~\cite{singh2017hide}. 

\begin{table}
  \centering
    \begin{tabular}{lcccc}
\toprule
Name		&Res 	&Size	&Scene		&Real\\
\midrule
Sintel~\cite{Butler:ECCV:2012}		&0.45			&1064	&In/Outdoor	&N          \\
ETH3D~\cite{schoeps2017cvpr}		&0.46			&27		&Campus		&Y          \\
KITTI-15~\cite{Menze2015CVPR}		&0.47			&200	&Driving	&Y	        \\
Sceneflow~\cite{MIFDB16}	&0.52			&30k	&All		&N          \\
\midrule
Middlebury~\cite{scharstein2014high}	&6.00			&23		&Indoor		&Y          \\
HR-VS (Ours)		&5.07			&780	&Driving	&N  \\
HR-RS (Ours)		&4.65			&120&Driving	&Y          \\ 
\bottomrule
\end{tabular}
  \caption{Summary of datasets for stereo matching, where the first group contains low-resolution datasets and the second group contains high-resolution datasets. We can see: 1) There is a lack of large-scale high-resolution stereo dataset, especially for outdoor scenes. 2) There is also a lack of high-resolution stereo matching benchmark for driving scenario depth sensing. The datasets we proposed bridge these gaps.  Resolution (Res.) is shown in megapixel.}
\label{tab:dataset}%
\end{table}

\subsection{High-resolution datasets}
End-to-end high-resolution stereo matching requires high-resolution datasets to make it effective. However, as shown in Table \ref{tab:dataset}, there exist very few such datasets for both training and testing purpose. Middlebury-v3 \cite{scharstein2014high} is the only publicly available dataset for high-resolution stereo matching. However, it contains very few samples and there are no outdoor/driving scenes. To bridge this gap, we aggregate two datasets for high-resolution stereo matching on driving scenes. Synthetic HR-VS is collected for training high-resolution stereo models, while the high-res real stereo (HR-RS) dataset is collected to benchmark high-resolution stereo matching methods under real-world driving scenes.

\begin{figure}
\centering
\includegraphics[width=1\linewidth]{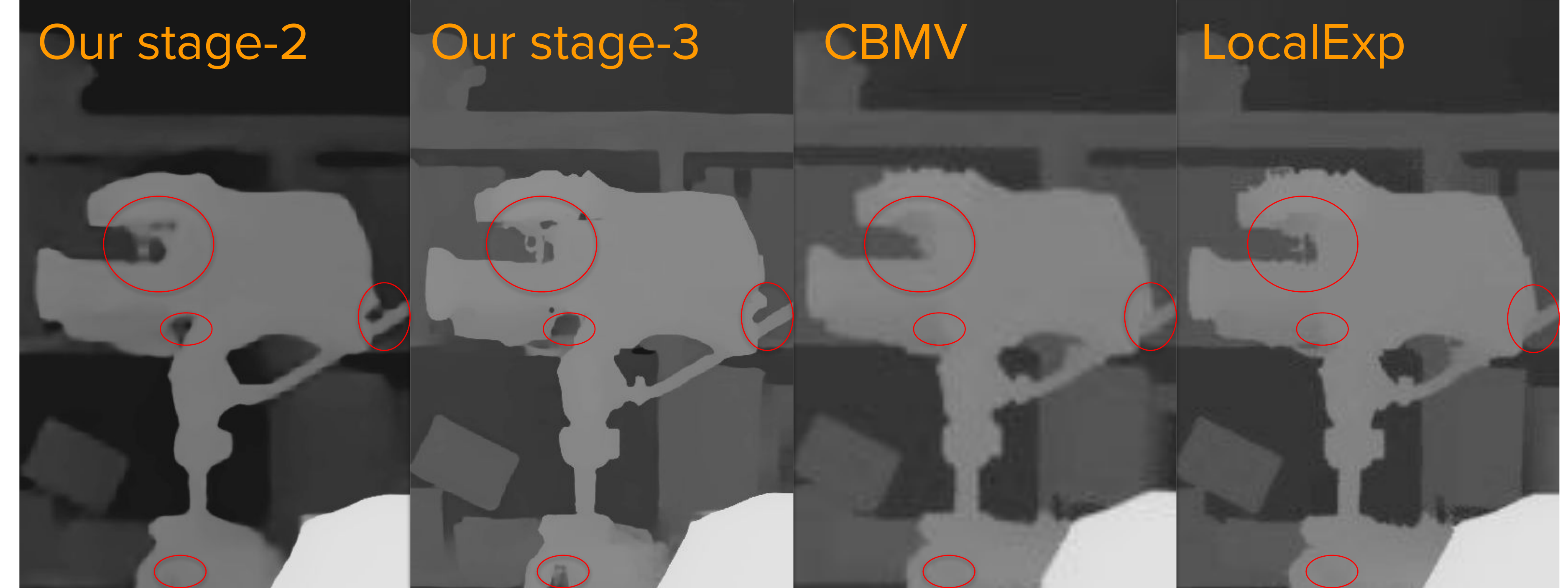}
\includegraphics[width=0.95\linewidth]{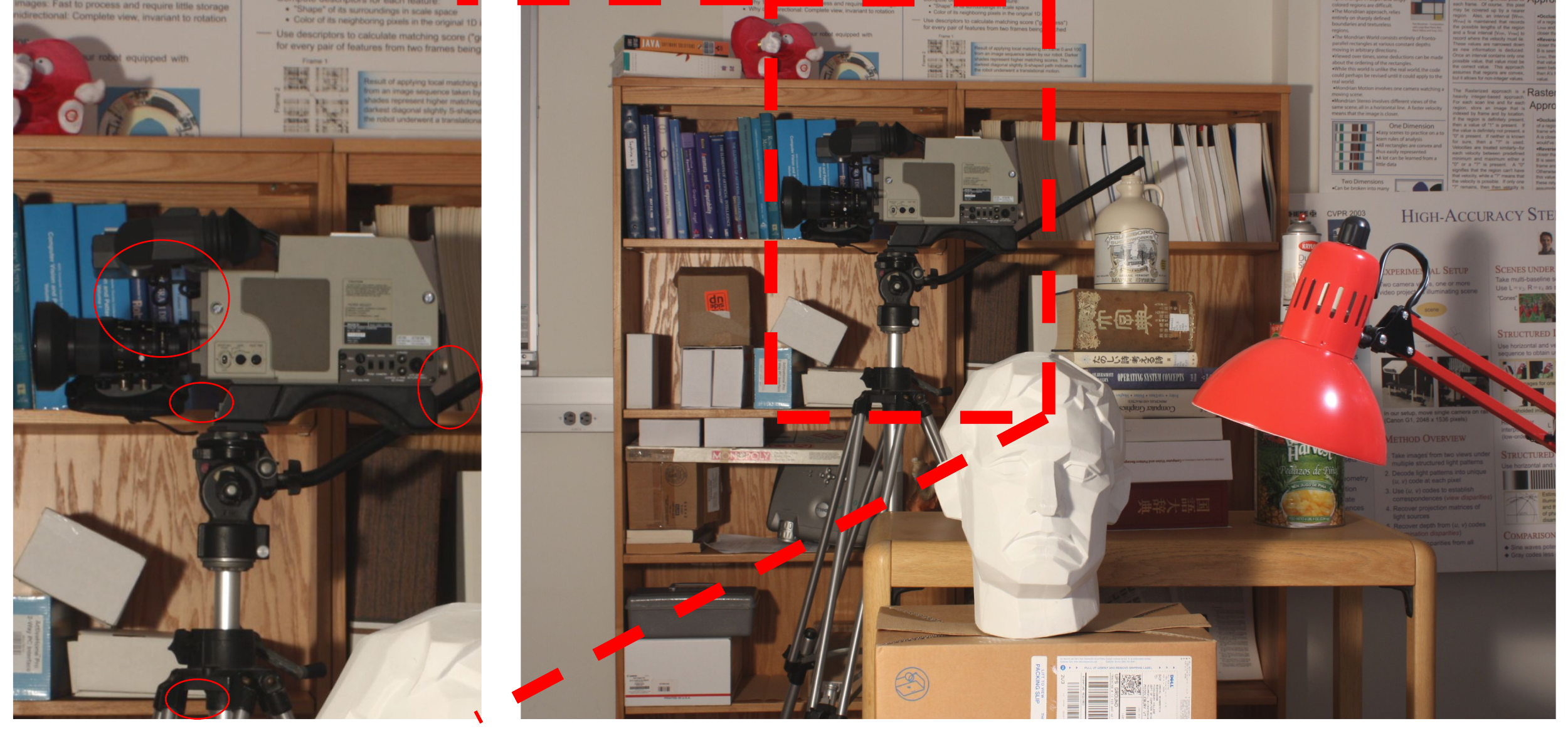}
\caption{Effectiveness of high-res on Middlebury test image ``Newkuba". LocalExp~\cite{taniai2018continuous} and CBMV\_ROB~\cite{batsos2018cbmv} take half-res inputs and rank 1st and 2nd on the bad-1.0 metric over all published methods. As shown in the circled regions, our method gives better details on thin structures.}
\vspace*{-0.1in}
\label{fig:comparison}
\end{figure}

{\bf High-res virtual stereo (HR-VS) dataset:}
HR-VS is collected using the open-sourced Carla simulator \cite{Dosovitskiy17}. Under $4$ weather conditions while maneuvering in Town01, we collected $780$ pairs of training data at $2056\times2464$ resolution. Camera baseline and focal length are set as $0.54$m and $3578$px respectively. Pixels with a depth greater than $200$m or a disparity greater than $768$px are removed to mimic the disparity distribution of real-world driving scenarios, resulting in a disparity range of $\lbrack9.66,768\rbrack$px and depth range of $\lbrack2.52,200 \rbrack$m. Sample images and ground-truth can be found in the supplements.

\begin{figure*}
\includegraphics[width=\linewidth]{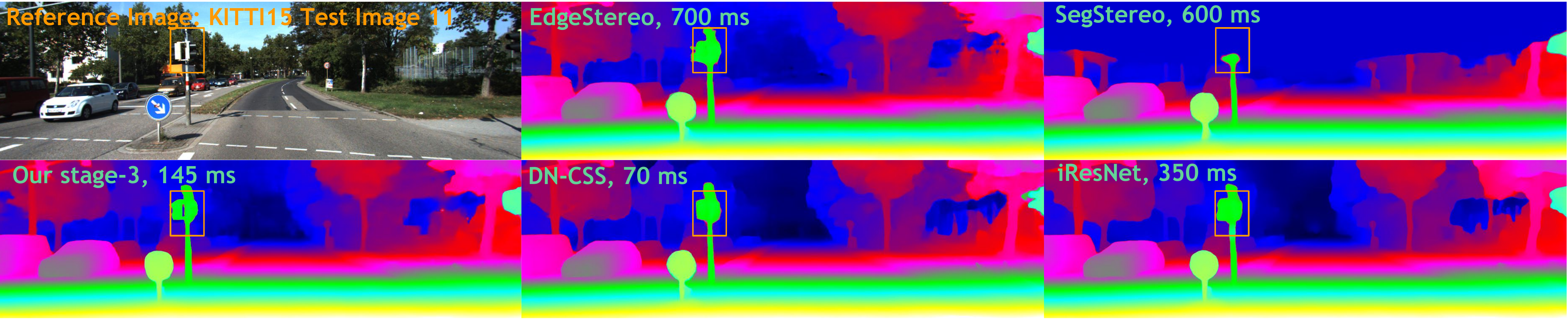}
\caption{
Qualitative results on KITTI-15 benchmark. As shown in the brown rectangle, our hierarchical stereo matching algorithm correctly finds the skinny structures and depth discontinuities while running faster than most SOTA (at $6.9$ FPS). }
\vspace*{-0.1in}
\label{middlebury-nkuba}
\end{figure*}

{\bf High-res real stereo (HR-RS) benchmark:}
HR-RS includes $33$ pairs of images and disparity ground-truths collected using high-resolution stereo cameras and LiDAR while driving in the urban scenes. Images are rectified and cropped to $1918\times2424$. LiDAR point clouds are projected to image planes and converted to disparities, resulting in a range of $[5.4,182.3]$px. We also manually removed point clouds on dynamic objects to reduce camera-LiDAR registration error.

{\bf Benchmark protocol:}
Since we are motivated by autonomous depth sensing, it is crucial to know the ideal sensing range under different driving speeds. Under the assumption of dry road condition and maximum deceleration of a car as shown in the supplements, we calculate the safe stopping distance for speed $v\in\{25,40,55\}$mph as $d\in\{25,60,115\}$m respectively. To ensure the safety of the driver and passengers, we are interested in making correct predictions within these stopping distances, i.e., when an object enters the stopping distance, we have to correctly perceive it. This gives us three sets of metrics with regard to driving speeds, which we refer to as short-range (0-25m), middle-range (25-60m) and long-range (60-115m) metrics. For each safe distance range, we use the same set of metrics in pixel space following \cite{scharstein2014high}.

\section{Experiments}
In this section, we evaluate our hierarchical stereo matching (HSM) network on public benchmarks including high-res Middlebury-v3 and low-res KITTI,  as well as our proposed high-resolution benchmark, i.e., HR-RS.

\begin{table*}
\centering
    \begin{tabular}{clcccccccc}
\toprule
Method          &time (s) &avgerr &rms & bad-4.0 & bad-2.0 & bad-1.0 &A99 &A95 &A90\\
\midrule
HSM-F3 (Ours)    & 0.51 (0.61)   &\underline{\bf{3.44}}$_1$ &\underline{\bf{13.4}}$_1$ & \bf{9.68}$_{7}$ &\bf{16.5}$_{16}$ &\bf{31.2}$_{28}$ &\underline{\bf{63.8}}$_{1}$ &\underline{\bf{17.6}}$_{1}$ &\underline{\bf{4.26}}$_{3}$\\
SGM\_ROB \cite{hirschmuller2008stereo}  
                                & 0.32          &14.2 &47.5 &19.1 &26.4 &38.6 &231  &97.5 &31.1\\
iResNet\_ROB \cite{Liang2018Learning}   
                                & 0.34 (0.42)   &6.56 &18.1 &22.1 &31.7 &45.9 &87.5 &36.2 &15.1\\
ELAS\_ROB \cite{geiger2010efficient}
                                & 0.48          &13.4 &34.9 &26.0 &34.6 &51.7 &152  &79.8 &38.8\\
PSMNet\_ROB \cite{chang2018pyramid} 
                                & 0.64          &8.78 &23.3 &29.2 &47.2 &67.3 &106  &43.4 &22.8\\
DN-CSS\_ROB \cite{ilg2018occlusions}
                                & 0.66          &5.48 &16.8 &19.6 &28.3 &41.3 &82.0 &25.6 &13.3\\
\midrule
LPS-H \cite{sinha2014efficient}
                                & 9.52          &19.7 &44.7 &23.3 &27.6 &38.8 &169  &108 &58.3\\
SPS \cite{legendre2017high} 
                                & 25.5          &24.8 &65.8 &23.6 &28.1 &36.9 &284  &172 &89.8\\
LPS-F \cite{sinha2014efficient}
                                & 25.8          &22.3 &54.1 &24.7 &28.8 &36.0 &219  &134 &70.2\\
MC-CNN-acrt \cite{zbontar2016stereo}     
                                & 150           &17.9 &55.0 &15.8 &19.1 &27.3 &261  &140 &56.6\\
CBMV\_ROB \cite{batsos2018cbmv}     
                                & 3946          &6.65 &27.7 &10.3 &13.3 &21.6 &134    &36.7  &8.41\\
LocalExp \cite{taniai2018continuous}     
                                & 881          &5.13 &21.1 &\underline{8.83} &\underline{11.7} &\underline{21.0} &109    &31.6  &5.27\\
\bottomrule
\end{tabular}
 \caption{Results on Middlebury-v3 benchmark where all pixels are evaluated. The subscript number shows the absolute rank among the benchmark, best results over the ``fast" group are bolded, and best results overall are underlined. While running at $510$ms/image on $6$-megapixel images, our method out-performed all algorithms running faster under $1$s/image by a large margin. Over the whole benchmark, we achieved $1$st place on avgerr, rms, and A99, and A95. 
  }
\label{tab:middlubury14}%
\end{table*}

\subsection{Setup}
{\bf Datasets:} We used $4$ publicly available datasets, including Middlebury-v3, KITTI-15, ETH3D and Sceneflow \cite{MIFDB16,Menze2015CVPR,scharstein2014high,schoeps2017cvpr}, as well as HR-VS dataset for training. Middlebury-v3 contains $10$ high-resolution training image pairs where each has variants with imperfect calibration, different exposures, and different lighting conditions, resulting in $60$ pairs in total. KITTI-15 contains $200$ low-resolution pairs and ETH3D contains $27$ low-resolution pairs with sparsely labeled ground-truth. Sceneflow contains around $30$k synthetic image pairs with dense disparity labels. And HR-VS contains $780$ training pairs.

{\bf Implementation:} 
 We implemented the HSM network using Pytorch. We train the model using Adam optimizer with a batch size of 24  on a machine with $4$ Tesla V100 GPUs, while setting the initial learning rate to $0.001$ and betas to $(0.9, 0.999)$. We train for $9$ epochs and then shrink the learning rate by $10$. 

During training, we augment Middlebury, KITTI-15, ETH3D~\cite{MIFDB16,Menze2015CVPR,scharstein2014high} and HR-VS to the same size as Sceneflow, resulting in around 170k training samples. We perform both symmetric and asymmetric augmentations on the fly. Asymmetric chromatic augmentations include randomly applying different brightness ($\lbrack 0.5,2 \rbrack$), gamma ($\lbrack 0.8,1.2 \rbrack$) and contrast ($\lbrack 0.8,1.2 \rbrack$) to target and inference images. We apply y-disparity augmentation by uniformly sample a delta y-translation ($\lbrack 0,2 \rbrack$px) and rotation($\lbrack 0,0.1 \rbrack\deg$) at a chance of $0.5$. We also apply asymmetric masking at a chance of $0.5$ by uniformly sample the width ($\lbrack 50,150 \rbrack$px) and height ($\lbrack 50,150 \rbrack$px) of the mask and randomly placing it on the image. Symmetric augmentations include scaling ($\lbrack 0.9,4.8 \rbrack$ for low-res images and $\lbrack 0.225,1.2\rbrack$ for high-res images) and randomly cropping to fix-sized $(576\times 768)$ patches. We set the number of disparities for searching as $768$px. 

At test time, we make predictions at the last $3$ scales from coarse to fine. With full-res input, these predictions are referred to as ``Our-F1", ``Our-F2" and ``Our-F3" respectively, where ``F'' indicates the input resolution and the following number indicates the stage. Similarly, with half resolution input we have ``Our-H2'' and ``Our-H3''; for quarter resolution inputs, we have ``Our-Q3''. When testing on Middlebury-v3 images, we set the disparity search range according to maximum disparities in the calibration files, while for 1.8-times upscaled KITTI-15 test images, we set disparity search range as 384. For HR-RS images, we set disparity search range as 512.


{\bf Metrics:} Different metrics are used for different benchmarks. On Middlebury-v3 \cite{scharstein2014high}, we divide official metrics into $3$ groups: 1) bad-4.0 (percentage of ``bad" pixels whose error is great than 4.0), bad-2.0 and bad-1.0, which tolerates the small deviations such as quantization error. 2) avgerr (average absolute error in pixels) and rms (root-mean-square disparity error in pixels), which also takes into account the subpixel accuracy. 3) A99 ($99\%$ error quantile in pixels), A95 and A90, which ignores large deviations while measuring the accuracy. On KITTI-15 \cite{Menze2015CVPR}, we use official metrics D1-all, D1-bg and D1-fg, which measure the percentage of outliers for all pixels, background pixels and foreground pixels respectively. While on HR-RS, we separate pixels to different depth ranges and use the same set of metrics as Middlebury-v3.

\subsection{Benchmark performance}

{\bf High-res Middlebury-v3:}
Because we could not submit multiple on-demand outputs to the online test server for Middlebury-v3, we evaluate only our full-res, full-pipeline model HSM-F3. We compare against two groups of published methods. The first group of methods includes those running under {\bf 1s/image}, which we refer to as the ``fast" group, and the second group of methods includes published SOTA that run slower. To compare with iResNet \cite{Liang2018Learning}, we also add the $13$ additional images as they did in training time, and in test time, we take the full resolution input images as input.

We first present qualitative results, including comparisons with SOTA methods in Fig.~\ref{fig:comparison} and  reconstruction of Middlebury test scene ``Plants" in Fig.~\ref{fig:middlebury-vis-plants}. The quantitative comparisons to SOTA are shown in Table \ref{tab:middlubury14}. Compared to MC-CNN-acrt on all-pixels, we reduced avgerr by $80.8\%$, rms by $75.6\%$ and bad-4.0 by $38.7\%$ while running $294.1$ times faster. Compared to CBMV\_ROB, we reduced avgerr by $48.3\%$, rms by $51.6\%$ and bad-4.0 by $6.0\%$ while running $7737.3$ times faster. 
We run $0.31$ times slower than iResNet\_ROB but reduced avgerr by $47.6\%$, rms by $26.0\%$ and bad-4.0 by $56.2\%$. Notice that in submission time, we used Tesla V100 on Amazon AWS, which gives us 510ms/image on average while iResNet used Titan Xp. To be fair, we measured running time for HSM-F3 and ``iResNet" on the same Titan X Pascal GPU, as shown in parenthesis. We are $31.4\%$ slower, but much more accurate. 


{\bf Low-res KITTI:}
Though our focus is not on low-res, we still evaluate on KITTI-15 and partition the SOTA algorithms into two groups: those that run under $200$ms and fits real-time needs\footnote{Since many autonomous robots employ sensors synced at 10fps, the deployment-ready version of a 200ms model is likely to fit real-time needs.}, while the other group is slower but more accurate. During training time, we excluded the ETH3D and Sceneflow datasets and finetuned with Middlebury-v3, KITTI-12 and KITTI-15.  At test time, we operate on image pairs up-sampled by a factor of $1.8$. We first show qualitative results in Figure \ref{middlebury-nkuba}. Then we refer to Table \ref{tab:kitti-2015} for quantitative results. We rank $1$st among all published methods while running $3.8$ times faster than ``EdgeStereo", which ranks $2$nd among the published methods.

\begin{table}
    \begin{tabular}{clccc}
\toprule
Method & D1-all  &D1-bg & D1-fg &time (ms) \\
\midrule
HSM-stage-3 (Ours) & \underline{\bf{2.14}}     & \underline{\bf{1.80}} & \bf{3.85}& 150\\ 
DN-CSS~\cite{ilg2018occlusions}  & 2.94          & 2.39 & 5.71  & 70 \\
DispNetC~\cite{MIFDB16}     & 4.34         & 4.32 & 4.41 & 60 \\
StereoNet~\cite{khamis2018stereonet}    & 4.83          & 4.30 & 7.45 & 20\\ 
\midrule
EdgeStereo~\cite{song2018edgestereo}      & 2.16       & 1.87 & \underline{3.61}  & 700\\
SegStereo~\cite{yang2018segstereo}       & 2.25    & 1.88 & 4.07 & 600 \\
PSMNet~\cite{chang2018pyramid}          & 2.32     & 1.86 & 4.62 & 410 \\ 
PDSNet~\cite{tulyakov2018practical}          & 2.58      & 2.29 & 4.05 & 500 \\

CRL~\cite{pang2017cascade}             & 2.67     & 2.48 & 3.59 & 470\\
iResNet~\cite{Liang2018Learning}         & 2.71      & 2.27 & 4.89  & 350\\ 
GCNet~\cite{kendall2017end}           & 2.87      & 2.21 & 6.16 & 900\\
\bottomrule
\end{tabular}
  \caption{Results on KITTI-15 benchmark where all pixels are evaluated and error metrics are shown in ($\%$). Best results over the ``real-time" group are bolded, and best results overall are underlined.}
\label{tab:kitti-2015}%
\end{table}

\subsection{Results on HR-RS}
Then we evaluate on HR-RS and compare against a subset of the previous arts under the protocol discussed in method section. ELAS~\cite{geiger2010efficient} is taken from the Robust Vision Challenge official package, iResNet~\cite{Liang2018Learning} is taken from their Github repository, and  we implemented two-pass SGBM2~\cite{hirschmuller2008stereo} using OpenCV (with SAD window size = 3, truncation value for pre-filter = 63, p1 = 216, p2 = 864, uniqueness ratio = 10, speckle window size = 100, speckle range = 32). The results from SGBM2 is also post-processed using weighted least square filter with default parameters. ``HSM-F2" means the model operates on full resolution images and make predictions at the second stage (2nd last scale). Results are shown in Table \ref{tab:argo-depth}.

{\bf Anytime on-demand:}
Cutting off HSM-F at the second stage (HSM-F2), bad-4.0 on all pixels increases by $10.3\%$ while being $1.46$x faster, which is still more accurate than ELAS, SGBM and iResNet. Same as iResNet-H, HSM-H3 uses half resolution image as input, but runs $4.54$x faster and produces 29.5\% lower bad-4.0. Halting earlier (HSM-H2) increases error
by 47.8\% but is 0.76x faster, which is still more accurate
than ELAS and SGM.

{\bf Long-range sensing:}
We analyze our methods for different distance ranges. Halting HSM-F earlier at the second stage (HSM-F2) only increases bad-4.0 on short-range pixels by 5.1\%, which suggests that high-res inputs might not help. However, on long-range pixels it increases bad-4.0 by 14.8\%. Interestingly, halting HSM-F earlier (HSM-F2) still produces more accurate long-range predictions than HSM-H3 (17.1\% versus 19.7\%). This is possibly because the features pyramid already encodes detailed information from high-res inputs, which helps to predict accurate long-range disparities.



\begin{table}
\centering
    \begin{tabular}{lccccc}
\toprule
&  &\multicolumn{4}{c}{bad-4.0 (\%)} \\
\cmidrule(lr){3-6}
Method & time (ms)& S &M &L &All\\
\midrule
HSM-F1
    & 91     &42.3&43.5&33.7&40.9\\
HSM-F2      
    & 175    &16.5&18.8&17.1&17.1\\
HSM-F3 
  &430           &{\bf15.7}&{\bf16.7}&{\bf14.9}&{\bf15.5}\\
\midrule
HSM-H2
    &42     &25.9&27.6&34.4&26.9\\
HSM-H3      
  &74       &18.9&18.9&19.7&18.2\\
\midrule
HSM-Q3      
  &29        &30.3&32.7&33.4&31.2\\
\midrule
ELAS-H~\cite{geiger2010efficient}   
    &464       &49.4&32.2&23.9&36.1\\
SGBM2-Q~\cite{hirschmuller2008stereo}   &1321       &50.8&27.3&19.5&32.8  \\
iResNet-H~\cite{Liang2018Learning} 
    &410      &32.1&24.4&22.8&25.8\\
\bottomrule
\end{tabular}
  \caption{Results on HR-RS dataset. All pixels are evaluated at $3$ ranges. S: 0-25m, M: 25-60m, L: 60-115m. HSM achieves significant improvement in all metrics compared to the baselines.}
\label{tab:argo-depth}%
\vspace{-0.2cm}
\end{table}

\subsection{Diagnostics}
We perform a detailed ablation study to reveal the contribution of each individual component of our method. We follow the same training protocol as described in the experiment setup, but initialize the encoder weights using the same fine-tuned model. We train with a batch size of $8$ for $60$k iterations. The learning rate is down-scaled by $10$ for the last $10$k iterations. We then test on half-resolution additional Middlebury images except for "Shopvac", resulting in 24 test images in total. Quantitative results are shown in Table \ref{tab:diag}.

\begin{table}
    \begin{tabular}{lcccc}
\toprule
Method &avgerr & bad-1.0 & bad-2.0* &time (ms) \\
\midrule
Full-method	    &4.01	&{\bf46.93}	&{\bf26.88} & 97\\
\midrule
Cost-aggre.     &4.02 	&53.07	&31.03 & 98\\
- HR-VS         &4.02  &51.01 &30.24    &97\\
- ydisp         &4.28   &48.75 &28.98 & 98\\
- multi-scale   &4.20   &48.53  &28.83 & 97\\
- masking     &4.05   &48.89 & 28.20 & 97 \\
- VPP           &{\bf 3.88}   &47.99  &27.73 & 93\\
- A-Chrom.      &3.91   &46.96 &27.02 & 97\\
\bottomrule
\end{tabular}
 \caption{Diagnostic by removing each individual component. ``Cost-aggre." means replace feature volume fusion by cost volume aggregation. Results are ranked by bad-2.0.}
\label{tab:diag}%
\end{table}

{\bf Feature volume fusion:}
When aggregating information across pyramid scales, we made the design choice to fuse coarse-scale 4D {\bf feature} volumes instead of 3D {\bf cost} volume. Our intuition was that ``feature-volume fusion" tolerates incorrect coarse predictions since the final output does not directly depend on initial predictions. Also as shown in Table \ref{tab:diag}, we found that replacing ``feature-volume fusion" with ``cost-volume fusion" results in $13.1\%$ more bad pixels with an error greater than $1$px.

{\bf High-res synthetic data:}
After removing HR-VS dataset when training the network, the bad-1.0 metric increases by $8.7\%$, and the bad-2.0 metric increases by $11.1\%$, which shows the importance of adding synthetic high-res data when training a high-res stereo network.

{\bf y-disparity augmentation:}
y-disparity augmentation is an effective way of forcing the network to learn features robust to camera calibration error. As shown in Table \ref{tab:diag}, removing y-disparity augmentation increase bad-1.0 by $7.2\%$. 

{\bf Multi-scale loss:}
Multi-scale loss regularizes the network by forcing it to learn multiple prediction tasks, while also helps gradients propagate through multiple scales. We found that removing Multi-scale loss increase bad-2.0 by $6.8\%$. 

{\bf Asymmetric masking:}
Asymmetric masking is a powerful tool that encourages the network to learn to deal with occlusions. After removing ``asymmetric masking" data augmentation, bad-2.0 error increases by 4.9\%.  

{\bf Volumentric pyramid pooling}
Volumetric pyramid pooling effectively enlarges the receptive field of the decoder network, which is important when the input resolution is high. Removing VPP results in an increase of bad-2.0 error by 3.2\%. 

{\bf Asymmetric chromatic augmentation:}
We found removing asymmetric chromatic augmentation does not harm performance on Middlebury additional images. This is probably because we introduce extra noise into training pairs, which harms our predictions on normal images (with the same exposure/lighting). However, we found this technique is helpful in making the network robust to asymmetric imaging conditions.

\subsection{Robustness to adversarial conditions}

\begin{figure*}
\centering
\includegraphics[width=\linewidth]{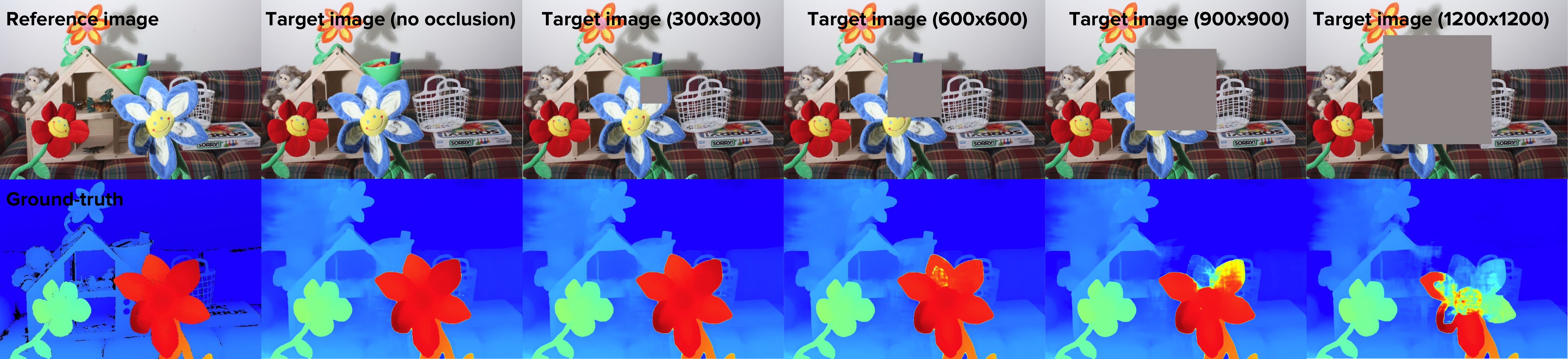}
\caption{Qualitative result of robustness to camera occlusion.}
\label{fig:occlusion2}
\end{figure*}

\begin{figure}
\centering
\includegraphics[width=\linewidth]{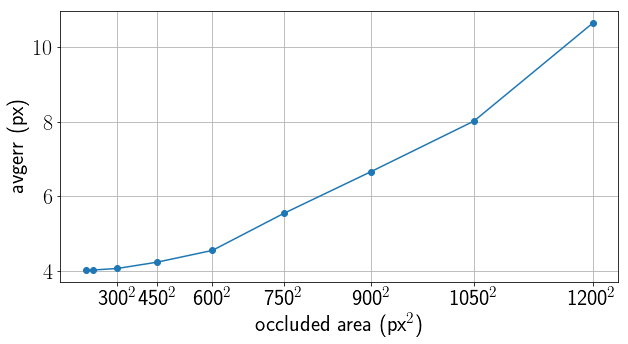}
\caption{Quantitative result of robustness to camera occlusion.}
\label{fig:occlusion1}
\end{figure}

\textbf{Robustness to camera occlusion}
In real-world autonomous driving scenarios, it may occur that one of the stereo cameras is blurred by the raindrops or occluded by the snowflakes, which introduces challenges to stereo matching algorithms. To study the effect of occlusion to our model, we place a rectangular gray patch at the target image center while keeping the reference view unchanged. Predictions from our model are shown in Fig.~\ref{fig:occlusion2} and Fig.~\ref{fig:occlusion1}. We find that with an increased amount of occluded pixels, the average error grows slowly at the beginning (0.1\% from 0 to with 5625 occluded pixels), and faster towards the end. Our model is capable of handling small camera occlusions, possibly by inferring disparities based on monocular and context cues.

\textbf{Resilience to calibration error}
We also compare our HSM-H model against ELAS-H~\cite{geiger2010efficient} in terms of robustness to calibration error as shown in Tab.~\ref{tab:res-ydisp}. We find our model is more robust to calibration errors: when the camera becomes imperfectly calibrated, the average pixel error rate of HSM-H only increases 5.9\%, while that of ELAS-H increases by 29.6\%. The robustness of our model is due to the coarse-to-fine matching process where the calibration error is suppressed in the coarse scale. Our y-disparity augmentation is also helpful in that it forces the network to learn features robust to y-disparities.

\begin{table}
\centering
    \begin{tabular}{lccc}
\toprule
 &\multicolumn{3}{c}{avgerr (px)}\\
\cmidrule(lr){2-4} Method & perfect~$\downarrow$ & imperfect~$\downarrow$ &  increase (\%)~$\downarrow$ \\
\midrule
HSM-H	    & {\bf3.9} & {\bf4.1} &{\bf5.9}\\
ELAS-H~\cite{geiger2010efficient}      & 8.3 & 10.8 & 29.6\\
\bottomrule
\end{tabular}
  \caption{Resilience to calibration error on Middlebury additional images. For each of the $12$ image pairs, we evaluate separately on both its perfect and imperfect-calibrated version and compute the increase of error rate when the calibration becomes imperfect.}
\label{tab:res-ydisp}%
\end{table}

\begin{figure}
\centering
\includegraphics[width=0.470\linewidth]{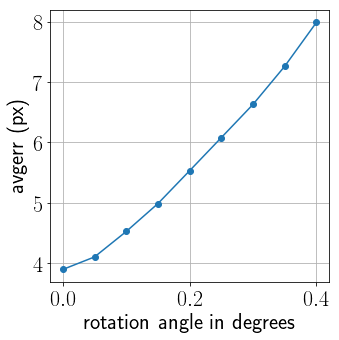}
\includegraphics[width=0.490\linewidth]{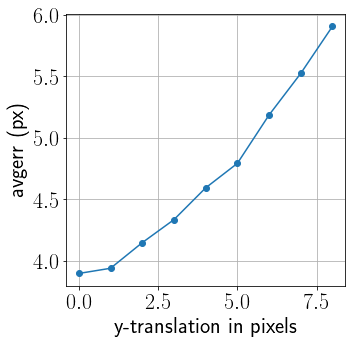}
\caption{Resilience to calibration error.}
\label{fig:y-disp-res}
\end{figure}

We then characterize the sensitivity of our model to calibration errors as follows: we select images with perfect calibration (12 in total) and 1) rotate the target view images around center for $\theta\in\{0,0.05,0.1,\dots,0.4\}$ degrees, as well as 2) translate along y-axis for $y_d\in\{0,0.5,1,\dots,4\}$ pixels, while keeping the reference view unchanged. The resulting avgerr to is shown in Fig.~\ref{fig:y-disp-res}.  We find that under a small rotation of 0.05 degrees (resulting 1.07 px y-disparity around the image border), our model's error rate only increases by 5.3\%, while under a large rotation of 0.4 degrees (resulting 8.6 px y-disparity around the image border), our model's error increases by 105\%, while still better than the results of ELAS-H on perfectly calibrated images (8.0 vs 8.3). Also under a small translation of 1px, our model's error rate almost does not increase (only rise by 1.0\%), while under a large translation of 8 px, our model's error increases by 51.5\%.

\textbf{Robustness to change of exposure}
We further compare our HSM-H model to ELAS-H in terms of robustness to exposure change and robustness to lighting changes as shown in Tab.~\ref{tab:res-e}, where we found our model to be more robust: when exposure changes, the average error of HSM-H only increases 8.2\%, while that of ELAS-H increases by 44.2\%. The robustness of our model also shows the effectiveness of asymmetric chromatic augmentation, which forces the model to learn features robust to chromatic changes.

\begin{figure}
\centering
\includegraphics[width=\linewidth]{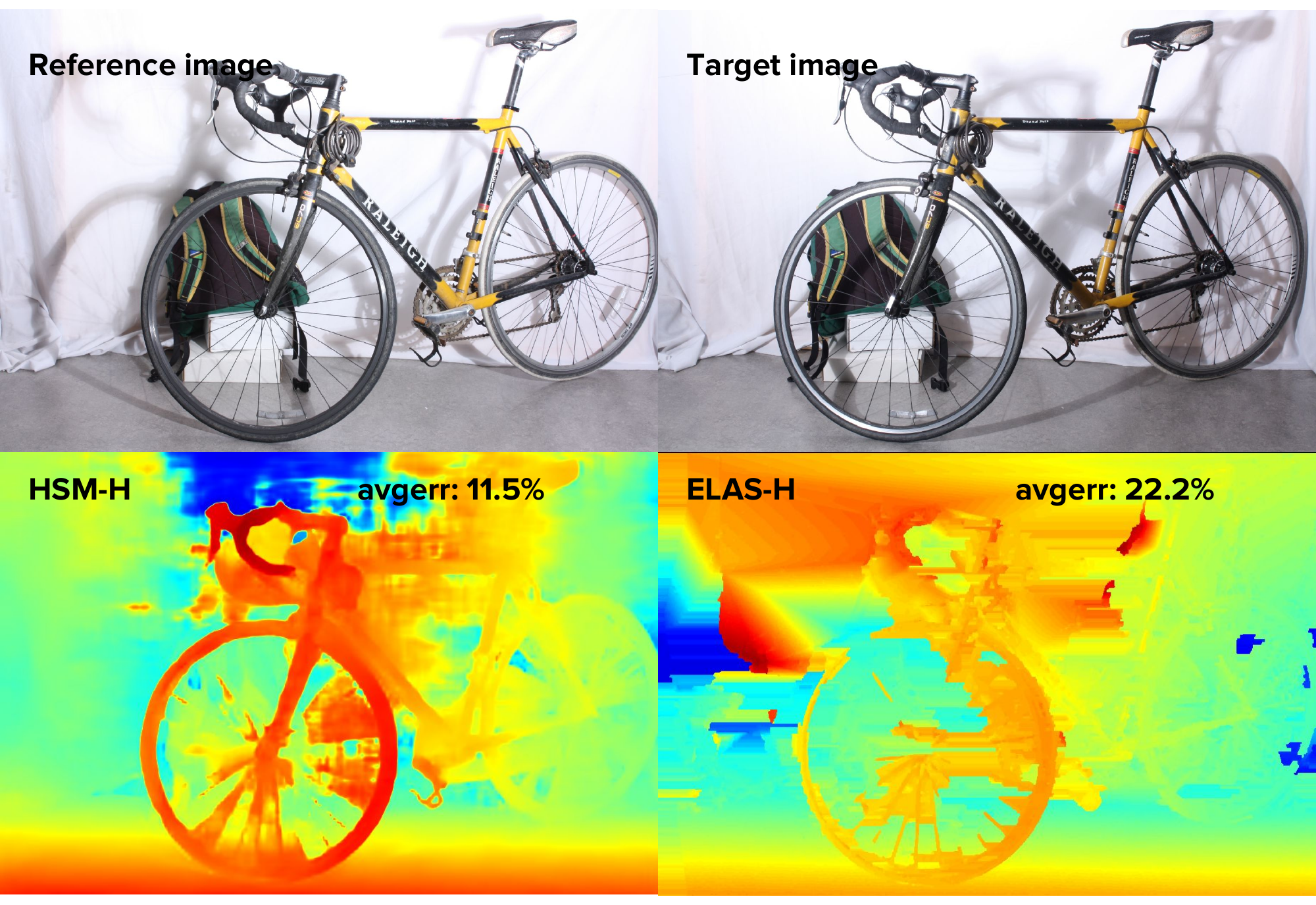}
\caption{A failure case of our model on "Bicycle1" with changed lighting condition. Notice that change of lighting condition introduces different shading to reference and target images, and both HSM and ELAS cannot handle that.}
\label{fig:sup-res-l}
\end{figure}

\textbf{Robustness to change of lighting}
We finally test HSM-H and ELAS-H in terms of robustness to lighting changes, and results are shown in Tab.~\ref{tab:res-l}. We found that when lighting changes, the average error of both HSM-H and ELAS-H increase a lot. We also include a failure case for both our model and ELAS as in Fig~\ref{fig:sup-res-l}.

\begin{table}
\centering
    \begin{tabular}{lccc}
\toprule
 &\multicolumn{3}{c}{avgerr (px)}\\
\cmidrule(lr){2-4} Method & normal~$\downarrow$ & E~$\downarrow$ &  increase (\%)~$\downarrow$\\
\midrule
HSM-H	    & {\bf 4.0} & {\bf4.3} &{\bf8.2}\\
ELAS-H~\cite{geiger2010efficient}      & 9.5 & 13.7& 44.2\\
\bottomrule
\end{tabular}
  \caption{Resilience to exposure change on Middlebury14 additional images. For each of the $24$ image pairs, we evaluate separately on both its original and changed exposure (E) version, and calculate the increase of error rate when the exposure changes.}
\label{tab:res-e}%
\end{table}

\begin{table}
\centering
  \begin{tabular}{lccc}
\toprule
 &\multicolumn{3}{c}{avgerr (px)}\\
\cmidrule(lr){2-4} Method & normal~$\downarrow$& L~$\downarrow$ &  increase (\%)~$\downarrow$\\
\midrule
HSM-H	    & {\bf 4.0} & {\bf9.3} &{\bf131.1}\\
ELAS-H~\cite{geiger2010efficient}      & 9.5& 23.6& 148.4\\
\bottomrule
\end{tabular}
  \caption{Resilience to lighting change on Middlebury additional images. For each of the $24$ image pairs, we evaluate separately on both its original and changed lighting (L) version, and calculate the increase of error rate when the exposure changes.}
\label{tab:res-l}%
\end{table}

\section*{Conclusion}
With the help of hierarchical designs, high-resolution synthetic dataset and asymmetric augmentation techniques, our model achieves SOTA performance on Middlebury-v3 and KITTI-15 while running significantly faster than prior arts. We are also able to perform on-demand disparity estimation at different scales, making possible accurate depth prediction of close-by objects in real-time.

{\bf Acknowledgements:} This work was supported by the CMU Argo AI Center for Autonomous Vehicle Research.

{\small
\bibliographystyle{ieee}
\bibliography{egbib}
}

\end{document}